\documentclass[10pt,twocolumn,letterpaper]{article}

\usepackage[pagenumbers]{anonymous_conf} %

\usepackage{microtype}

\renewcommand{\paragraph}[1]{\vspace{.5em}\noindent\textbf{#1}}

\definecolor{anonymous_conf_blue}{rgb}{0.21,0.49,0.74}
\usepackage[pagebackref,breaklinks,colorlinks,allcolors=anonymous_conf_blue]{hyperref}

 \usepackage{placeins}
\usepackage{subcaption}
\usepackage{relsize}
\usepackage{tabularx}
\usepackage{graphicx}
\usepackage{caption}
\usepackage[accsupp]{axessibility}  %
\newsavebox\CBox

\usepackage{xcolor}
\usepackage{overpic}
\definecolor{darkgreen}{rgb}{0.0, 0.5, 0.0}
\definecolor{pink}{rgb}{1,0.75,0.8}
\usepackage{colortbl}

\graphicspath{{images/}}
\usepackage{amsmath}
\usepackage{dsfont}

\usepackage{algpseudocode}
\usepackage[ruled,vlined]{algorithm2e}

\newcommand{\figref}[1]{Fig.~\ref{#1}}
\newcommand{\tabref}[1]{Tab.~\ref{#1}}
\newcommand{\secref}[1]{Sec.~\ref{#1}}

\usepackage{booktabs}
\usepackage{multirow}
\usepackage{overpic}
\usepackage{enumitem}
\usepackage{xcolor}
\usepackage{subcaption}
\usepackage{amsfonts}
\usepackage{amsmath}
\usepackage{float}
\usepackage{booktabs}
\usepackage{wrapfig}

\usepackage[utf8]{inputenc} %
\usepackage[T1]{fontenc}    %
\usepackage{hyperref}       %
\usepackage{url}            %
\usepackage{amsfonts}       %
\usepackage{nicefrac}       %

\def\confName{anonymous\_conf}
\def\confYear{}

\title{PartAgent: Agentic 3D Part Segmentation via Expert Integration and Closed-Loop Refinement}

\author{%
Yang Cao$^{1}$ \quad
Zhe Zhu$^{2}$ \quad
Yiheng Zhang$^{1}$ \quad
Zhengqiang Zhang$^{3}$ \\
Haoyuan Li$^{4}$ \quad
Le Wan$^{2}$ \quad
Dan Xu$^{1}$\textsuperscript{*}
  \vspace{0.1cm}
  \\
  $^{1}$Hong Kong University of Science and Technology \quad
  $^{2}$Tencent  \\
  $^{3}$Hong Kong Polytechnic University \quad
  $^{4}$Sun Yat-sen University\\[3pt]{\normalsize \enspace\href{https://yangcaoai.github.io/PartAgent_ProjectPage/}{\nolinkurl{yangcaoai.github.io/PartAgent_ProjectPage/}}}
}

\begin{document}
\clearcaptionsetup{sub}
\twocolumn[{
\renewcommand\twocolumn[1][]{#1}
\maketitle
\begin{center}
    \vspace{-12pt}
    \includegraphics[width=\textwidth]{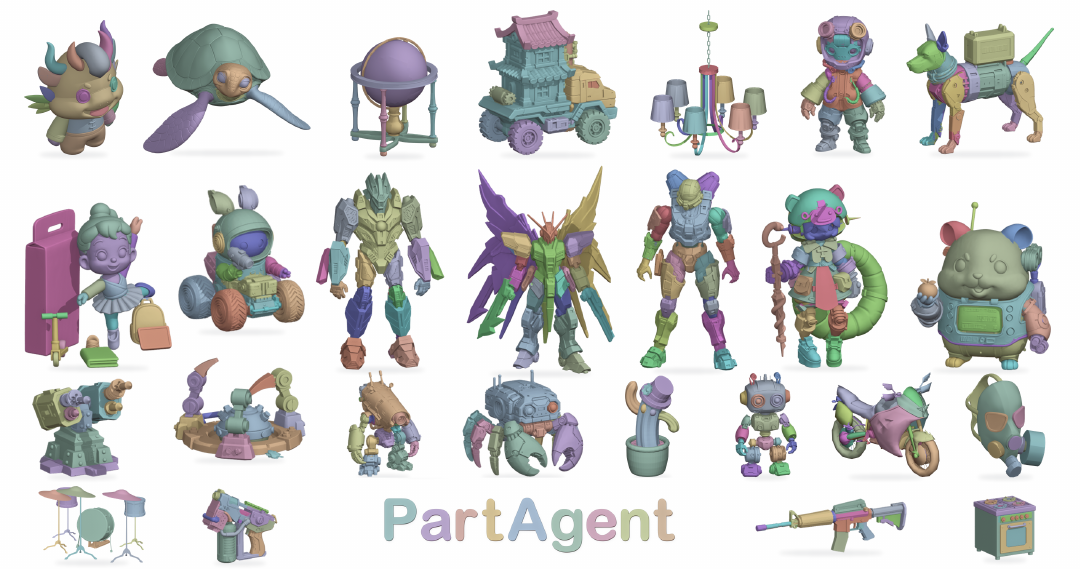}
    \captionsetup{hypcap=false}
    \captionof{figure}{\textbf{PartAgent on diverse 3D assets.}
We show PartAgent's segmentation results on meshes from existing benchmarks and meshes generated by Hunyuan3D~\cite{lai2025hunyuan3d25highfidelity3d}, ranging from characters and vehicles to everyday objects. Different colors indicate segmented parts.}
    \label{fig:hunyuan_showcase}
\end{center}
}]
\begingroup
\renewcommand{\thefootnote}{\fnsymbol{footnote}}
\footnotetext[1]{Corresponding author.}
\endgroup
\begin{abstract}
Segmenting a 3D object into semantically meaningful parts requires
joint reasoning about geometric structure and semantic context.
Pretrained 3D experts provide spatially grounded region proposals,
while 2D experts provide semantic cues for identifying object parts.
The challenge is to integrate these complementary capabilities into
a framework that can both produce meaningful part segmentations and
progressively refine them.
We present \textbf{PartAgent}, a training-free agentic framework for
3D part segmentation built upon two complementary modules:
(i)~Structural--Semantic Expert Integration combines regions predicted by the 3D expert with multiview semantic evidence from the 2D expert through image-to-surface correspondences, constructing compositional part partitions at multiple
granularities.
(ii)~Observation-Driven Closed-Loop Partition Refinement observes the
current partition and its supporting evidence to select targeted
inspection, evidence repair, composition, or reselection actions.
It verifies proposed changes and uses the updated state and action
outcomes to guide subsequent decisions.
On PartObjaverse-Tiny-WT, PartAgent achieves 68.41 mIoU, outperforming the strongest competing method by 8.53 points. Across 3DCoMPaT200, PartNeXt, PartNetE, and PartObjaverse-Tiny, PartAgent achieves the highest mIoU on each dataset, averaging 69.36 across the four datasets, 7.28 points above the highest mean among competing methods.

\end{abstract}

\section{Introduction}
\label{sec:introduction}

3D part segmentation aims to partition a 3D object into meaningful parts~\cite{ma2025p3sam,zhu2026partsam,liu2025partfield,yang2024sampart3d,ma2025find3d,dong2025copart,dong2024interactive3d}, which is a fundamental task in 3D understanding~\cite{mo2019partnet,cao2023coda,qi2016pointnet,qi2017pointnetplusplus,cao2026vggtdet,wang2025n3d}.
Doing so requires both geometric cues about the object's structure
and semantic context to determine which regions belong together.
3D experts~\cite{liu2025partfield,ma2025p3sam} propose spatially grounded
regions, while image-based approaches~\cite{liu2023partslip,tang2024samesh,yang2024sampart3d}
provide visual grouping and semantic context across views.

\begin{figure}[!t]
    \centering
    \includegraphics[width=\linewidth]{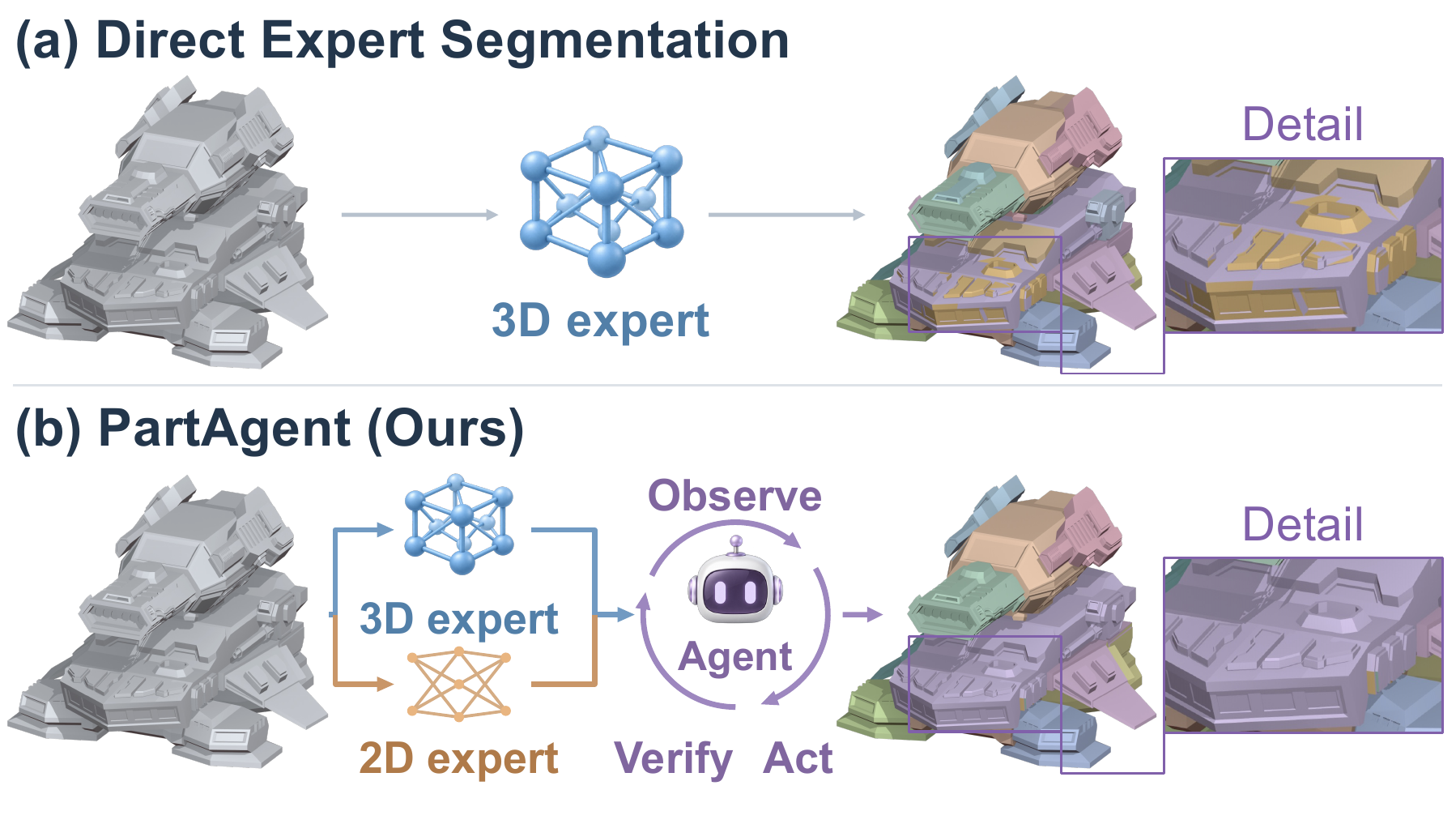}
    \caption{\textbf{From a 3D expert to an agentic framework.}
(a) A 3D expert directly segments the object into parts.
(b) PartAgent integrates predictions from frozen 2D and 3D experts
and iteratively refines the segmentation through
an observation-driven loop.
    }
    \vspace{-0.4cm}
    \label{fig:teaser}
\end{figure}

Turning these complementary cues into meaningful parts remains
challenging: (i)~A 3D proposal may cover only a fragment of a component or
span several components, while semantic predictions from different views
may disagree about part boundaries or identities. Constructing a coherent partition therefore requires jointly considering geometric and semantic evidence to determine how regions should be grouped into parts.
(ii)~Even after integration, the initial partition may still be imperfect and require further inspection and revision.
To address both challenges, we present \textbf{PartAgent}~(\figref{fig:teaser}), a training-free agentic framework that combines geometric and semantic evidence from pretrained 3D and 2D experts to construct meaningful part partitions,
and then iteratively refines the segmentation by selecting targeted actions based on current observations.

To construct meaningful initial partitions, our key intuition is to treat spatially grounded regions as building blocks and use semantic evidence to guide how they are refined and grouped into parts.
We propose \emph{Structural--Semantic Expert Integration}, which maps multiview semantic predictions onto the original mesh through image-to-surface correspondences and uses the mapped evidence to refine 3D expert regions.
It then composes these regions into candidate partitions at multiple granularities, selects an initial partition, and retains alternative candidates for later revision.

Further refinement requires deciding where to inspect, what to revise, and whether a proposed change improves the result.
We propose \emph{Observation-Driven Closed-Loop Partition Refinement}, which uses a vision--language model (VLM) to examine projected partitions, geometry renders, and semantic predictions and select a targeted action. The agent can inspect a region, repair 2D segmentation masks, revise part groupings, or select an alternative candidate. The agent verifies each proposed update before adoption. Both acceptance and rejection guide later actions.

On PartObjaverse-Tiny-WT~\cite{ma2025p3sam}, PartAgent
achieves 68.41 mIoU, outperforming the strongest competing
method by \textbf{8.53 points}. Across 3DCoMPaT200~\cite{ahmed2024dcompat}, PartNeXt~\cite{wang2026partnext}, PartNetE~\cite{liu2023partslip}, and
PartObjaverse-Tiny~\cite{yang2024sampart3d}, PartAgent achieves the highest mIoU on
each dataset, with a four-dataset mean of 69.36, exceeding
the strongest competing method's mean by \textbf{7.28 points}. \figref{fig:hunyuan_showcase}  shows PartAgent segmenting diverse, complex 3D assets drawn from existing benchmarks or generated with Hunyuan3D~\cite{lai2025hunyuan3d25highfidelity3d} into parts.

Our contributions are threefold:
\begin{itemize}
    \item \textbf{A Training-Free Agentic Framework for 3D Part Segmentation.}
    We integrate frozen 2D and 3D experts to construct meaningful part
    partitions and iteratively refine them through VLM-guided decisions.

    \item \textbf{Structural--Semantic Expert Integration.}
    We use multiview semantic evidence to refine 3D expert regions
    and compose candidate partitions at multiple granularities.

\item \textbf{Observation-Driven Closed-Loop Partition Refinement.}
We introduce an observation-driven loop that adaptively refines
both part groupings and their supporting semantic evidence.
Verification determines which updates to adopt and provides
feedback for subsequent action selection.
\end{itemize}

\section{Related Work}
\label{sec:related_work}

\noindent\textbf{3D part representations and segmentation.}
Point and mesh networks encode geometry using dynamic
graphs~\cite{wang2019dgcnn}, mesh-edge convolutions~\cite{hanocka2019meshcnn},
and attention~\cite{zhao2021pointtransformer,wu2024pointtransformerv3}.
PartNet~\cite{mo2019partnet} provides hierarchical part annotations.
PartField~\cite{liu2025partfield} clusters part-aware features, while
P3-SAM~\cite{ma2025p3sam} predicts native 3D masks with automatic
selection and merging. PartSAM~\cite{zhu2026partsam} uses native 3D
supervision for scalable, promptable part segmentation.
SegviGen~\cite{li2026segvigen} adapts a native 3D generative model
to color-code object parts. PartGen~\cite{chen2024partgen} combines multiview
part segmentation and completion with 3D reconstruction.
PartAgent builds candidate segmentations from 2D and 3D predictions and chooses refinement actions based on what it observes.

\begin{figure*}[t]
    \centering
    \includegraphics[width=\textwidth]{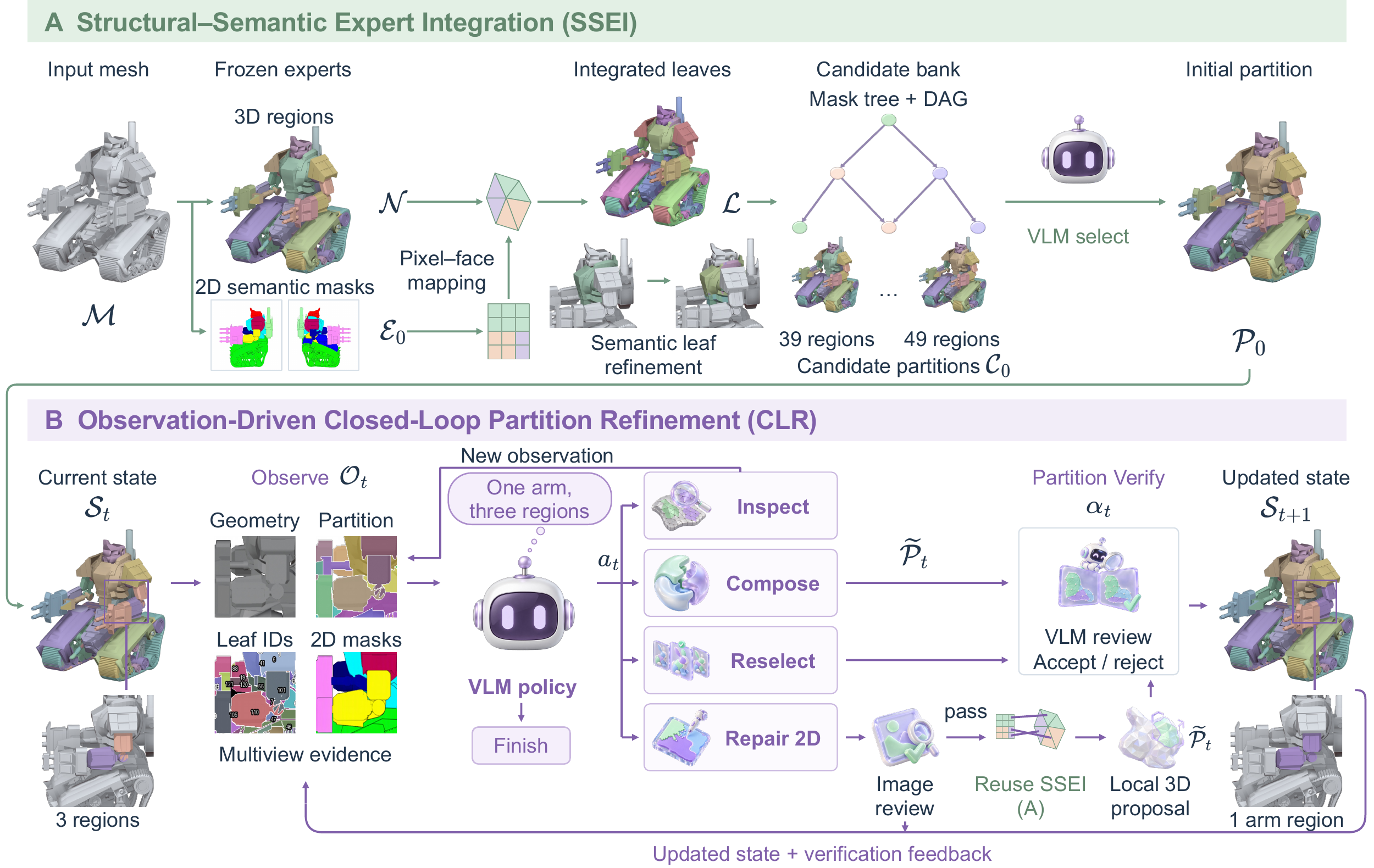}
    \caption{\textbf{Overview of PartAgent.}
(A) Structural--Semantic Expert Integration (SSEI) uses multiview
semantic evidence to refine 3D expert regions and combine them into
candidate segmentations at multiple granularities, then selects an
initial segmentation $\mathcal{P}_0$.
(B) Observation-Driven Closed-Loop Partition Refinement (CLR) uses a
VLM to examine the current segmentation and its supporting evidence
and choose targeted actions. Proposed segmentation updates are
verified before adoption, and feedback guides subsequent actions.}
\vspace{-0.4cm}
    \label{fig:partagent}
\end{figure*}

\noindent\textbf{Multiview transfer for 3D segmentation.}
Transformers and self-supervised learning~\cite{vaswani2017attention,dosovitskiy2021vit,caron2021dino}
enable pretrained models to extract visual features and interpret images~\cite{radford2021clip,li2022glip,oquab2024dinov2,yan2025pgocc,li2026thinking,Qwen3-VL,Qwen2.5-VL}.
Promptable models generate segmentation masks~\cite{kirillov2023segmentanything,carion2026sam}.
PartSLIP~\cite{liu2023partslip} lifts language-guided detections to point clouds.
SAMesh~\cite{tang2024samesh} links masks across mesh views, while
SAMPart3D~\cite{yang2024sampart3d} distills scale-conditioned features using multiview semantic labels.
S2AM3D~\cite{su2026s2am3d} combines 2D priors and 3D supervision
for consistent features and controllable segmentation granularity.
For scenes, OpenScene~\cite{peng2023openscene} and
OpenMask3D~\cite{takmaz2023openmask3d} transfer language-aligned
features to points and instance masks, respectively.
SAMPro3D~\cite{xu2025sampro3d} uses shared 3D prompts across views.
PartAgent's SSEI uses multiview semantics to refine and group 3D expert
regions, then selects one complete segmentation and retains alternatives for later revision.

\noindent\textbf{Agentic perception and refinement.}
Agentic methods act on intermediate observations~\cite{ag25ijcai,kong2026harnessir}.
ReAct~\cite{yao2023react} interleaves reasoning and action, and
Reflexion~\cite{shinn2023reflexion} retains feedback without updating weights.
Visual Programming~\cite{gupta2023visprog} and ViperGPT~\cite{suris2023vipergpt}
compose pretrained modules into executable programs.
VASA~\cite{wang2026vasa} inspects and edits masks for open ad-hoc
2D segmentation. Agent3D-Zero~\cite{zhang2024agent3dzero}
selects viewpoints for 3D understanding. A3R~\cite{li2026a3r}
learns to gather evidence sequentially for instruction-conditioned
affordance grounding in Gaussian scenes.
For 3D part segmentation, PartAgent uses observations and verification
feedback to repair multiview 2D masks, regroup 3D regions, or choose another
complete segmentation from its candidates.

\providecommand{\abltbd}{\textemdash}
\providecommand{\ablna}{\textit{n/a}}
\newcolumntype{Y}{>{\raggedright\arraybackslash}X}

\newcommand{\PartAgentDesignAblationTables}{%
\begin{table*}[t]
\noindent\begin{minipage}[t]{\columnwidth}
    \caption{\textbf{Structural--Semantic Expert Integration.}
    We ablate Semantic Leaf Refinement (Leaf ref.) and Mask DAG in
    \figref{fig:partagent}A  without CLR. The final row is
    Full SSEI (SSEI only in \tabref{tab:ablation_core_contributions}). Both components improve segmentation performance.}
    \label{tab:ablation_expert_integration}
\end{minipage}\hfill
\begin{minipage}[t]{\columnwidth}
    \caption{\textbf{Improving different 3D experts.}
    Entries show Original $\rightarrow$ PartAgent, with gains in points
in parentheses. 
The 2D expert and VLM are fixed. PartAgent significantly improves segmentation performance with all three tested
3D experts.}
    \label{tab:ablation_3d_experts}
\end{minipage}
\par\noindent
\begin{minipage}[t]{\columnwidth}
    \vspace{0pt}
    \centering\normalsize
    \setlength{\tabcolsep}{0pt}
    \renewcommand{\arraystretch}{1.18}
    \begin{tabular}{@{}*{4}{>{\centering\arraybackslash}p{.25\linewidth}}@{}}
        \toprule
        Leaf ref. & Mask DAG & Partition $\uparrow$ & Mask $\uparrow$ \\
        \midrule
        -- & $\checkmark$ & 63.93 & 65.31 \\
        $\checkmark$ & -- & 61.52 & 63.87 \\
        $\checkmark$ & $\checkmark$ & \textbf{65.11} & \textbf{67.60} \\
        \bottomrule
    \end{tabular}
\end{minipage}\hfill
\begin{minipage}[t]{\columnwidth}
    \vspace{0pt}
    \centering\normalsize
    \setlength{\tabcolsep}{0pt}
    \renewcommand{\arraystretch}{1.18}
    \begin{tabular}{@{}>{\raggedright\arraybackslash}p{.28\linewidth}*{2}{>{\centering\arraybackslash}p{.36\linewidth}}@{}}
        \toprule
        3D expert & Partition $\uparrow$ & Mask $\uparrow$ \\
        \midrule
        P3-SAM~\cite{ma2025p3sam}
        & \mbox{\small 58.51\,$\scriptstyle\rightarrow$\,\textbf{65.98}\,{\footnotesize (+7.47)}}
        & \mbox{\small 58.51\,$\scriptstyle\rightarrow$\,\textbf{68.94}\,{\footnotesize (+10.43)}} \\
        PartSAM~\cite{zhu2026partsam}
        & \mbox{\small 52.30\,$\scriptstyle\rightarrow$\,\textbf{61.13}\,{\footnotesize (+8.83)}}
        & \mbox{\small 52.30\,$\scriptstyle\rightarrow$\,\textbf{64.12}\,{\footnotesize (+11.82)}} \\
        PartField~\cite{liu2025partfield}
        & \mbox{\small 50.54\,$\scriptstyle\rightarrow$\,\textbf{55.53}\,{\footnotesize (+4.99)}}
        & \mbox{\small 56.95\,$\scriptstyle\rightarrow$\,\textbf{59.30}\,{\footnotesize (+2.35)}} \\
        \bottomrule
    \end{tabular}
\end{minipage}
\par\vspace{-6pt}
\end{table*}
}

\section{Method}
\label{sec:method}

\begin{figure*}[!t]
    \centering

\includegraphics[width=\textwidth]{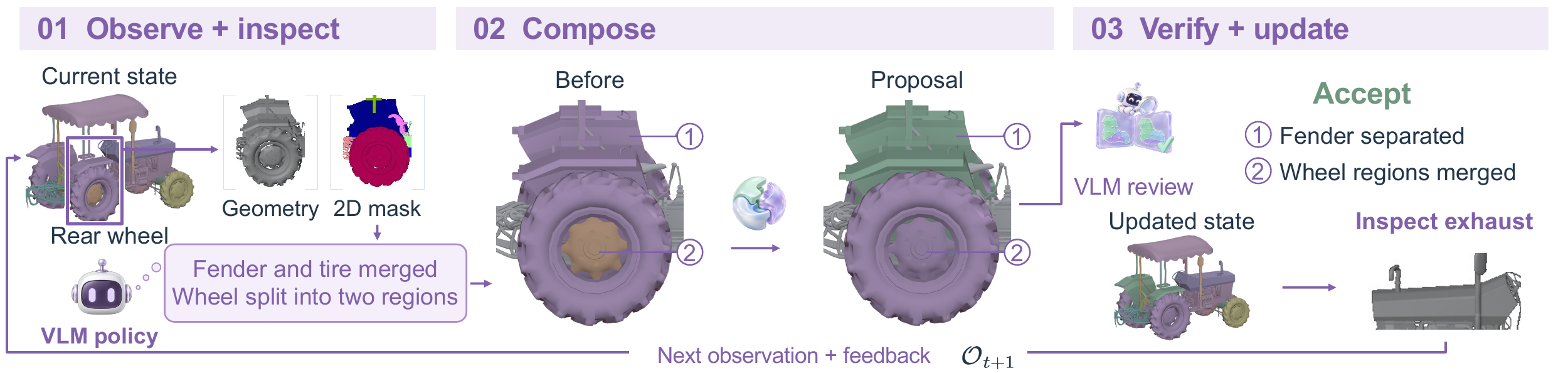}
\caption{\textbf{Observation-driven closed-loop refinement.}
Left: The agent inspects the rear wheel and identifies incorrect
part groupings.
Middle: It regroups existing 3D regions to separate the fender
from the tire and merge the two regions belonging to the same wheel.
Right: After the proposal passes multiview verification, the updated state and feedback guide inspection of the exhaust.
}
    \vspace{-0.4cm}
    \label{fig:closed_loop_example}
\end{figure*}

\subsection{Overview}
\label{sec:method_overview}

Given a triangular mesh $\mathcal{M}=(\mathcal{V},\mathcal{F})$,
we seek a partition $\mathcal{P}=\{P_1,\ldots,P_K\}$ whose regions
are disjoint and cover the input surface.
Here, $\mathcal V$ and $\mathcal F$ are the vertex and triangular-face sets.
Each $P_i\subseteq\mathcal F$ contains the faces of one of the $K$ parts. 
PartAgent combines predictions from frozen 3D and 2D experts,
and uses a vision--language model (VLM) for reasoning and verification.
We refer to the regions predicted by the 3D expert as \emph{3D expert regions}.
\figref{fig:partagent} summarizes the two key modules.
\emph{Structural--Semantic Expert Integration} (SSEI) uses multiview
semantic evidence to refine the 3D expert regions. The resulting regions,
called integrated leaves, are grouped into complete candidate partitions
from which SSEI selects an initial partition $\mathcal{P}_0$ (\secref{sec:expert_integration}).
\emph{Observation-Driven Closed-Loop Partition Refinement} (CLR)
observes the current partition and its supporting evidence, selects a
targeted action, and uses the outcome and verification feedback to guide
the next decision (\secref{sec:closed_loop}).
Each region ID refers to a set of mesh faces, and a complete partition
specifies the segmentation of the whole surface.

\subsection{Structural--Semantic Expert Integration}
\label{sec:expert_integration}

A 3D expert region may capture only a fragment of a component or span
several components. Image predictions provide semantic context, but
do not directly define a consistent surface partition.
We use image evidence to refine 3D expert regions, then construct candidate segmentations at different granularities, as shown
in \figref{fig:partagent}A.

\paragraph{Shared multiview evidence.}
We render the input mesh from multiple viewpoints to obtain images $I_v$
and retain the camera parameters and rasterized pixel--face maps,
\begin{equation}
    \phi_v:\Omega_v\rightarrow\mathcal{F}\cup\{\bot\},
\end{equation}
where $\Omega_v$ is the image domain and $\bot$ denotes background.
The VLM examines the rendered views together to propose and review a
part list describing each possible part's category and which visible
regions it should include or exclude.
Each entry in the part list receives a distinct color shared across views.
We keep this part list and color scheme fixed. The 2D expert then produces a color-coded part mask
for each target render using the other views as context.
We instruct the 2D expert to preserve the target render's silhouette,
visible surfaces, and pose.
These predictions form
$\mathcal{E}_0$ in \figref{fig:partagent}A.
We decode the mask colors and combine evidence across views by semantic category.

\paragraph{Semantic leaf refinement.}
The 2D masks may contain errors, so we weight their evidence before
using it to refine 3D expert regions.
Let $n_{vf}$ count the pixels projecting to face $f$ in view $v$,
and $n_{vfc}$ those assigned to category $c$.
We accumulate a score for each category as
\begin{equation}
\label{eq:surface_support}
    s_{fc}=\sum_v w_{vfc},\qquad
    w_{vfc}=\frac{n_{vfc}}{n_{vf}}\,r(n_{vf})\,\eta_{vf},
\end{equation}
with $w_{vfc}=0$ when face $f$ is not visible in view $v$.
Here, $r(n_{vf})$ is computed from the face's projected pixel count,
giving less weight to evidence based on only a few pixels.
The weight $\eta_{vf}$ is computed from the angle between the face normal
and the viewing direction, giving less weight when the face is viewed
from the side. A face receives a category only if the total score is
positive, the top category is unique, and the top two scores meet the margin condition:
\begin{equation}
\label{eq:fusion_margin}
    \frac{s_{f(1)}-s_{f(2)}}{\sum_c s_{fc}}\geq\tau,
\end{equation}
where $s_{f(1)}$ and $s_{f(2)}$ are the two largest category scores and
$\tau$ is the confidence-margin threshold.
We also propagate category support across adjacent faces.

\begin{figure*}[t]
    \centering
    \includegraphics[width=\textwidth]{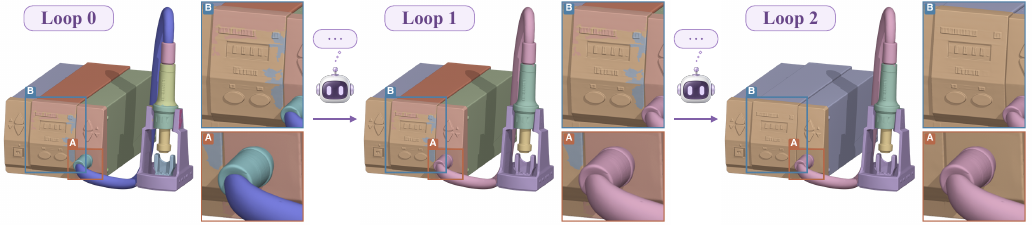}
    \caption{\textbf{Progressive improvement through closed-loop refinement.}
    Loop 0 denotes the SSEI initialization before closed-loop refinement. Loop 1 groups the cable and its connector into one part. Loop 2 merges fragments within the enclosure and within the control panel.
    Loop numbers denote agent decision rounds. Insets show the same boxed regions at each loop.}
    \vspace{-0.1cm}
    \label{fig:clr_evolution}
\end{figure*}

\begin{table*}[t]
    \centering
    \small
    \setlength{\tabcolsep}{3pt}
    \renewcommand{\arraystretch}{1.12}
    \caption{\textbf{Comparison on PartObjaverse-Tiny-WT}~\cite{ma2025p3sam}.
    All six methods are evaluated on the same watertight meshes.
    For each category, we report Mask Match mIoU averaged over its objects (\%, $\uparrow$).
    Overall is the average across all objects.
    Bold marks the highest score. Parentheses show the gain over the best baseline.}
    \label{tab:sota_watertight200_categories}
    \label{tab:sota_watertight200}
    \begin{tabular*}{\textwidth}{@{\extracolsep{\fill}}l*{9}{c}@{}}
        \toprule
        Method & Human & Animals & Daily & Build. & Trans. & Plants & Food & Elec. & Overall \\
        \midrule
        Find3D~\cite{ma2025find3d} & 24.58 & 23.19 & 24.00 & 14.69 & 12.72 & 23.06 & 29.96 & 18.49 & 19.90 \\
        SAMPart3D~\cite{yang2024sampart3d} & 60.39 & 70.73 & 63.03 & 50.59 & 51.78 & 65.18 & 70.53 & 58.33 & 59.53 \\
        PartField~\cite{liu2025partfield} & 61.54 & 65.83 & 69.19 & 47.83 & 48.79 & 65.36 & 73.31 & 62.77 & 59.88 \\
        P3-SAM~\cite{ma2025p3sam} & 59.65 & 57.68 & 61.60 & 42.99 & 49.65 & 56.56 & 64.47 & 55.22 & 54.85 \\
        PartSAM~\cite{zhu2026partsam} & 54.93 & 61.10 & 56.47 & 40.97 & 48.74 & 61.85 & 49.38 & 53.80 & 53.12 \\
        \midrule
        \textbf{PartAgent} & \textbf{70.97} & \textbf{71.00} & \textbf{70.92} & \textbf{56.40} & \textbf{62.93} & \textbf{75.40} & \textbf{78.00} & \textbf{71.61} & \textbf{68.41~(+8.53)} \\
        \bottomrule
    \end{tabular*}
    \vspace{-6pt}
\end{table*}

Let $\mathcal{N}=\{N_1,\ldots,N_J\}$ be the partition output by the 3D expert.
We highlight each 3D region in the rendered views and ask the VLM to
suggest a category label. We use these region-level cues together with
the fused image evidence to refine each $N_j$, separating faces assigned to different
categories and grouping faces assigned to the same category within that region.
These refined regions form the integrated leaves
$\mathcal{L}=\{L_1,\ldots,L_m\}$, which satisfy
\begin{equation}
\label{eq:structural_leaves}
    L_\ell\subseteq N_{\pi(\ell)},\qquad
    \bigcup_{\ell:\pi(\ell)=j}L_\ell=N_j,
\end{equation}
where $\pi(\ell)\in\{1,\ldots,J\}$ indexes the 3D expert region containing that leaf.
Thus, semantic evidence can subdivide 3D expert regions while each original
region can still be recovered as a union of leaves. A shared category
does not automatically merge distinct 3D expert regions.
The next step composes these leaves into candidate parts.

\paragraph{Compositional candidate partitions.}
Leaves supply boundaries, but several leaves may belong to one part
or a larger assembly. To retain these alternatives, we construct the
`Mask tree + DAG' in \figref{fig:partagent}A.
We build a tree that records which leaves belong to each 3D expert region,
then add larger semantic groups made from whole 3D expert regions.
The VLM uses descriptions and supporting evidence for the leaves and
existing groups to propose additional part groupings as a directed
acyclic graph (DAG).

Let $\mathcal{G}_0$ contain the selectable leaf, 3D expert region, and
semantic group nodes. Each node $g\in\mathcal{G}_0$ denotes a mask
\begin{equation}
\label{eq:group_union}
    M_g=\bigcup_{\ell\in D(g)}L_\ell,
\end{equation}
where $D(g)$ contains its descendant leaf indices and is $\{\ell\}$
for leaf node $\ell$.
An initial candidate partition selects a set of nodes
$\mathcal{A}\subseteq\mathcal{G}_0$ with
\begin{equation}
\label{eq:valid_partition}
    \bigcup_{g\in\mathcal{A}}M_g=\mathcal{F},\qquad
    M_g\cap M_h=\varnothing\quad(g\neq h),
\end{equation}
where the disjointness condition applies to nodes in $\mathcal A$.
PartAgent automatically constructs a finite set $\mathcal{C}_0$ of
candidate segmentations at multiple granularities using part candidates
from the tree and DAG. Each segmentation covers the entire object with
non-overlapping parts.

\paragraph{Selecting the initial partition.}
The VLM compares candidate segmentations from
$\mathcal{C}_0$ by examining their projections alongside the original
renders and semantic evidence. It checks whether parts are complete,
whether the evidence supports their grouping, and
whether the segmentation uses an appropriate level of detail across
the object. The selected partition becomes $\mathcal{P}_0$.
Other candidates remain available for visual inspection and reselection during later refinement loops as the agent gathers new observations.

\begin{table*}[t]
\noindent\begin{minipage}[t]{0.635\textwidth}
\vspace{0pt}
    \centering
    \small
    \setlength{\tabcolsep}{0pt}
    \renewcommand{\arraystretch}{1.12}
    \begin{minipage}[t][4\baselineskip][t]{\linewidth}
    \caption{\textbf{Comparison across multiple datasets.}
    Object-averaged Mask Match mIoU (\%,~$\uparrow$).
    Avg. is the mean of the four dataset scores. Parentheses show the gain
    over the best baseline. `3DCoM' denotes 3DCoMPaT200.
    `Part-T' denotes PartObjaverse-Tiny.}
    \label{tab:sota_benchmarks}
    \label{tab:sota_nonwatertight200}
    \label{tab:sota_nonwatertight200_categories} %
    \label{tab:sota_multidataset600} %
    \end{minipage}\par
\begin{tabularx}{\linewidth}{@{}l
    *{2}{>{\centering\arraybackslash}X}
    @{\hspace{3pt}}
    *{3}{>{\centering\arraybackslash}X}@{}}
        \toprule
        Method & \makebox[\linewidth][c]{3DCoM~\cite{ahmed2024dcompat}}
        & \makebox[\linewidth][c]{PartNeXt~\cite{wang2026partnext}}
        & \makebox[\linewidth][c]{PartNetE~\cite{liu2023partslip}}
        & \makebox[\linewidth][c]{Part-T~\cite{yang2024sampart3d}} & Avg. \\
        \midrule
        Find3D~\cite{ma2025find3d} & 32.21 & 33.59 & 22.54 & 21.47 & 27.45 \\
        SAMPart3D~\cite{yang2024sampart3d} & 55.16 & 59.26 & 54.49 & 54.33 & 55.81 \\
        PartField~\cite{liu2025partfield} & 60.70 & 66.01 & 63.32 & 58.27 & 62.08 \\
        P3-SAM~\cite{ma2025p3sam} & 51.80 & 59.79 & 59.63 & 55.22 & 56.61 \\
        PartSAM~\cite{zhu2026partsam} & 57.34 & 62.23 & 54.97 & 65.58 & 60.03 \\
        \midrule
        \textbf{PartAgent} & \textbf{67.82} & \textbf{72.42} & \textbf{71.43} & \textbf{65.76} & \textbf{69.36~(+7.28)} \\
        \bottomrule
    \end{tabularx}
\end{minipage}\hfill
\begin{minipage}[t]{0.345\textwidth}
\vspace{0pt}
    \centering
    \small
    \setlength{\tabcolsep}{2.0pt}
    \renewcommand{\arraystretch}{1.56}
    \begin{minipage}[t][4\baselineskip][t]{\linewidth}
    \caption{\textbf{SSEI and CLR ablations.}
    Partition and Mask denote Partition Match mIoU and Mask Match
    mIoU, respectively (\%, $\uparrow$).}
    \label{tab:ablation_core_contributions}
    \end{minipage}\par
    \begin{tabularx}{\linewidth}{@{}>{\raggedright\arraybackslash}Xcccc@{}}
        \toprule
        Variant & SSEI & CLR & Partition & Mask \\
        \midrule
        3D Expert only & -- & -- & 58.51 & 58.51 \\
        3D Expert + CLR & -- & $\checkmark$ & 62.72 & 64.37 \\
        SSEI only & $\checkmark$ & -- & 65.11 & 67.60 \\
        \midrule
        \textbf{Full PartAgent} & $\checkmark$ & $\checkmark$ & \textbf{65.98} & \textbf{68.94} \\
        \bottomrule
    \end{tabularx}
\end{minipage}
\vspace{-6pt}
\end{table*}

\subsection{Observation-Driven Closed-Loop Refinement}
\label{sec:closed_loop}

Refinement can require different actions. An ambiguous region may need
closer inspection, inaccurate image evidence may need repair, and an
incorrect grouping may be corrected by regrouping leaves or selecting
an existing candidate. The VLM in \figref{fig:partagent}B uses observations
to choose the action, its spatial target, and when to finish.
Each action outcome guides the next decision.
In \figref{fig:closed_loop_example}, the agent inspects the tractor and
proposes separating its fender from the tire and merging the two regions
belonging to the same wheel by regrouping existing leaves. After the proposal is accepted, the agent inspects the exhaust.

\paragraph{Observe and decide.}
At iteration $t$, the agent maintains
\begin{equation}
\label{eq:agent_state}
    \mathcal{S}_t=(\mathcal{P}_t,\mathcal{C}_t,
                  \mathcal{L},\mathcal{E}_t,\mathcal{H}_t),
\end{equation}
which contains the current partition, candidate bank, fixed integrated
leaves, current image evidence, and action/verification history.
The `Observe' panels in \figref{fig:partagent}B show the original renders,
the current partition, leaf IDs, and 2D masks in the same views.
The agent also receives the part list, candidate descriptions, and each
leaf's face count, 3D center and spatial bounds, and visibility across views.
Leaf IDs identify the same face sets across views.
The VLM identifies the observed problem, the views showing it,
and the expected improvement before selecting an action:
\begin{equation}
\label{eq:action_policy}
    a_t=\mu_\theta(\mathcal{O}_t,\mathcal{H}_t),
\end{equation}
where $\mathcal{O}_t$ contains the images and accompanying part,
candidate, and region information,
and $\mu_\theta$ is a frozen VLM with fixed pretrained parameters $\theta$.
It chooses one of the following actions and specifies the target and
any additional instructions needed to carry out the selected action:
\begin{itemize}[leftmargin=*,itemsep=1pt,topsep=2pt]
    \item \textbf{Inspect} requests a crop of an existing view or
    projections of selected candidates. The returned images provide
    additional visual evidence for the next decision.
    \item \textbf{Compose} regroups leaves to merge fragments of a part or split an overmerged region along leaf boundaries.
    \item \textbf{Repair 2D} selects a view and affected leaves,
    and specifies how to repair the mask. The 2D expert revises the mask using the original renders, fixed part list, and color scheme.
    \item \textbf{Reselect} proposes an existing complete candidate
    after visual inspection, without creating new masks.
    \item \textbf{Finish} terminates when the observations do not
    support a further action to improve the segmentation.
\end{itemize}
For example, unclear part structure or uncertainty about an alternative
candidate can prompt Inspect before a change is proposed. When a
fragmented or incorrectly merged part can be corrected by regrouping
existing leaves, the agent can choose Compose. If an inspected candidate
appears to offer a more plausible segmentation, it can choose Reselect.
Repair 2D is reserved for visible errors in the 2D masks that
regrouping or reselection alone cannot address. Previous outcomes help
the agent reconsider unsuccessful proposals.

\paragraph{From 2D repairs to local 3D updates.}
The VLM first checks the repaired 2D mask against the original render,
previous mask, and other views. If the VLM accepts the repaired mask,
PartAgent combines it with
the masks from the other views and reuses SSEI's evidence integration
(\figref{fig:partagent}A) to produce a proposed segmentation
$\widetilde{\mathcal P}_t$. Changes are restricted to the area selected
for repair, while the remaining segmentation stays unchanged.
The proposal then undergoes partition verification against the original
renders before it can replace the current segmentation.

\begin{figure*}[t]
    \centering
    \includegraphics[width=.96\textwidth,trim=0 1bp 0 3bp,clip]{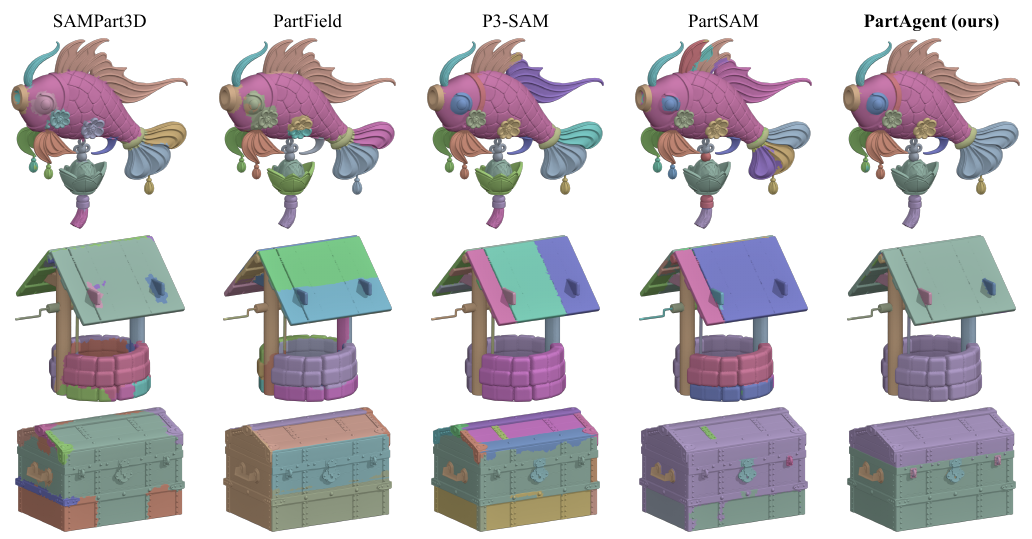}
    \caption{\textbf{Qualitative comparison.}
    Comparison with SAMPart3D~\cite{yang2024sampart3d}, PartField~\cite{liu2025partfield},
    P3-SAM~\cite{ma2025p3sam}, and PartSAM~\cite{zhu2026partsam}.
    On the Hunyuan3D fish, PartAgent keeps the dorsal and tail fins as one part each.
    On the well, PartAgent separates the roof fittings and segments the roof surface and stone wall as one part each.
    On the chest, PartAgent separates the lid from the body and keeps the handle and latch distinct.
    }
    \label{fig:qualitative_sota}
\end{figure*}

\PartAgentDesignAblationTables

\paragraph{Verifying changes and guiding the next action.}
A separate VLM call compares projections of the proposed segmentation
$\widetilde{\mathcal P}_t$ and the current partition against the original
renders (\figref{fig:partagent}B). Let $\alpha_t=1$ when the verifier
accepts the proposal as a meaningful improvement without major new
errors, and $\alpha_t=0$ otherwise. The current partition is updated as
\begin{equation}
\label{eq:verified_update}
    \mathcal{P}_{t+1}=
    \begin{cases}
        \widetilde{\mathcal{P}}_t, & \alpha_t=1,\\
        \mathcal{P}_t, & \alpha_t=0.
    \end{cases}
\end{equation}

The action, outcome, and reason for acceptance or rejection are recorded
in $\mathcal H_{t+1}$. The agent then observes the resulting segmentation
and image evidence, using feedback from both accepted and rejected
proposals to choose its next action. At termination, the current
partition is returned as the final segmentation of the input object.

\section{Experiments}
\label{sec:experiments}

\subsection{Experimental Setup}
\label{sec:implementation_details}

\paragraph{Datasets and evaluation.}
We evaluate all 200 objects in each of the PartObjaverse-Tiny~\cite{yang2024sampart3d} and PartObjaverse-Tiny-WT~\cite{ma2025p3sam} benchmarks.
To keep API costs manageable on the larger PartNeXt~\cite{wang2026partnext}, 3DCoMPaT200~\cite{ahmed2024dcompat}, and PartNetE~\cite{liu2023partslip} benchmarks, we select 500 objects per dataset through category-balanced random sampling.
We evaluate all methods on the same fixed subsets using their officially released code and pretrained models.
For the same reason, all ablations use the same fixed set of 100 meshes randomly sampled from PartObjaverse-Tiny-WT.
We will release all the evaluation data and splits.
Following PartField~\cite{liu2025partfield},
our main comparisons report \emph{Mask Match} mIoU.
For each object, we match each ground-truth part mask to the predicted
candidate mask with the highest IoU and average these IoUs. 
For each object, we average scores within each annotation level, then
across the available levels. Dataset scores average these object scores. 
Avg. in \tabref{tab:sota_benchmarks} is the mean of the four
dataset-level scores.
For ablations, we also report \emph{Partition Match} mIoU. For each object, we compute mIoU separately for each complete candidate partition using only its masks and report the highest value.
The two metrics complement each other.
\emph{Mask Match} evaluates individual part masks, while \emph{Partition Match} evaluates complete candidate partitions.

\paragraph{Models and inference.}
We use frozen P3-SAM~\cite{ma2025p3sam} as the 3D expert and
GPT-Image-2.5 as the 2D expert for image segmentation
and repair. We use \texttt{gpt-6.1-sol}
for reasoning and verification. We render each object from four fixed viewpoints.
For PartNetE point clouds, we use point neighborhoods and pixel--point
correspondences in place of mesh adjacency and pixel--face mappings.
In each controlled comparison, we change only the tested component
and keep all other settings fixed. For each object, CLR action and VLM
comparisons share the initial segmentation, candidate partitions, and
multiview evidence from SSEI.

\subsection{Comparison with Previous Methods}
\label{sec:comparison}

\noindent\textbf{Watertight meshes.}
As shown in~\tabref{tab:sota_watertight200},
PartAgent achieves 68.41 mIoU
on PartObjaverse-Tiny-WT, outperforming the strongest
competing method (59.88) by \textbf{8.53 points}. 
PartAgent also ranks first in all eight object categories.
The largest margins over the best competing score in each category are 11.15 points for transportation, 10.04 for plants, and 9.43 for human models.
All methods use the same matching procedure to compare their predicted
candidate masks with the ground-truth part masks for each object.

\noindent\textbf{Multiple benchmarks.}
\tabref{tab:sota_benchmarks} reports mIoU averaged over objects
for each dataset. PartAgent achieves
the highest mIoU on all four datasets, with 67.82 on 3DCoMPaT200, 72.42 on PartNeXt, 71.43 on PartNetE,
and 65.76 on PartObjaverse-Tiny.
Its mean mIoU across the four datasets is 69.36, exceeding the highest mean among competing methods (62.08) by \textbf{7.28 points}.

\noindent\textbf{Qualitative comparison.}
\figref{fig:qualitative_sota} compares PartAgent with competing methods.  PartAgent keeps individual parts together while separating smaller attached components.

\subsection{Ablation Studies}
\label{sec:ablations}
We randomly sample 100 meshes from PartObjaverse-Tiny-WT and use the same set for all ablations. In each score pair below, we report Partition Match mIoU first, followed by the corresponding Mask Match mIoU.

\paragraph{Two complementary mechanisms.}
\tabref{tab:ablation_core_contributions} isolates Structural--Semantic
Expert Integration (SSEI) and Observation-Driven Closed-Loop Partition
Refinement (CLR), corresponding to \figref{fig:partagent}A/B.
`SSEI only' raises the scores of `3D Expert only'
from 58.51/58.51 to 65.11/67.60. Adding CLR yields
`Full PartAgent' at 65.98/68.94, a further gain of 0.87/1.34 points.
Starting CLR directly from the 3D expert's segmentation yields
62.72/64.37 with the same reasoning and verification VLM, a gain of
\textbf{4.21/5.86} points over the expert alone. SSEI provides stronger initial
candidates, and the two modules together achieve the best scores on
both metrics.

\paragraph{Components of Structural--Semantic Expert Integration~(SSEI).}
\tabref{tab:ablation_expert_integration} evaluates the components of SSEI
without CLR. Without Semantic Leaf Refinement, the original 3D regions serve as
leaves. Scores fall by 1.18/2.29 points compared with full SSEI. Removing Mask DAG causes larger
losses of 3.59/3.73 points. Semantic Leaf Refinement uses image evidence to separate faces assigned
to different part categories within a 3D expert region. Mask DAG adds
candidate groupings that can bring fragments from different 3D expert
regions together as one part.

\begin{figure}[t]
\begin{minipage}{\columnwidth}
    \centering
    \includegraphics[width=\columnwidth]{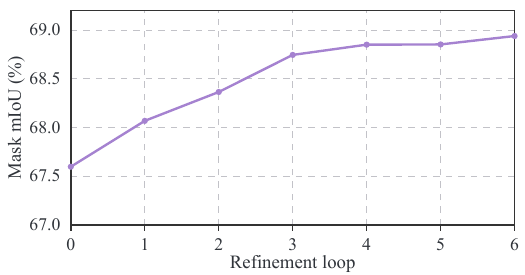}
    \captionsetup{skip=3pt}
    \captionof{figure}{\textbf{Performance across refinement loops.} }
    \label{fig:ablation_round_depth}
\end{minipage}
\vspace{-12pt}
\end{figure}

\paragraph{Improving different 3D experts.}
\tabref{tab:ablation_3d_experts} compares each 3D expert's original
predictions with PartAgent using the same expert. The 2D expert and
VLM remain fixed. PartAgent improves Partition Match/Mask Match mIoU
by \textbf{7.47/10.43} points with P3-SAM~\cite{ma2025p3sam}, \textbf{8.83/11.82} with PartSAM~\cite{zhu2026partsam}, and
\textbf{4.99/2.35} with PartField~\cite{liu2025partfield}. Our PartAgent improves results with each of the three
3D experts.

\paragraph{Choice of 2D expert.}
\tabref{tab:ablation_backbones} varies the 2D expert while retaining
P3-SAM and the default reasoning and verification VLM.
With GPT-Image-2.5, PartAgent scores 2.24/2.70 points higher than
with GPT-Image-2.
For VLM-prompted SAM3, the VLM provides positive and negative point
prompts for each rendered view of the input object. SAM3 uses these prompts and part names from the
fixed part list to produce 2D instance masks. PartAgent scores 0.36/1.50
points lower with SAM3 than with GPT-Image-2.5.

\paragraph{Dependence on the VLM.}
\tabref{tab:ablation_backbones} compares four VLMs for reasoning and
verification in CLR with the same SSEI initialization and prompts.
PartAgent performs best with GPT-6.1-sol on Partition Match (65.98)
and Mask Match (68.94). GPT-5.6 ranks second on Partition Match,
0.04 points ahead of GPT-6. GPT-6 ranks second on Mask Match,
0.15 points behind GPT-6.1-sol. Qwen3.5-9B ranks fourth on both metrics.
Across the tested 2D experts and refinement VLMs, PartAgent outperforms P3-SAM alone on both Partition Match and Mask Match mIoU (\tabref{tab:ablation_3d_experts}). These results show that PartAgent remains effective with different models for 2D mask generation and CLR.

\begin{table}[t]
\begin{minipage}{\columnwidth}
    \centering
    \normalsize
    \setlength{\tabcolsep}{0pt}
    \renewcommand{\arraystretch}{1.055}
    \captionof{table}{\textbf{2D expert and VLM substitutions.} P3-SAM is fixed. Each variant changes only the
    indicated component. }
    \label{tab:ablation_backbones}
    \begin{tabular}{@{}>{\raggedright\arraybackslash}p{.58\linewidth}*{2}{>{\centering\arraybackslash}p{.21\linewidth}}@{}}
        \toprule
        Component / variant & Partition & Mask \\
        \midrule
        \multicolumn{3}{@{}l}{\textit{2D expert}} \\
        GPT-Image-2.5  & \textbf{65.98} & \textbf{68.94} \\
        GPT-Image-2 & 63.74 & 66.24 \\
        VLM prompts + SAM3~\cite{carion2026sam} & 65.61 & 67.44 \\
        \midrule
        \multicolumn{3}{@{}l}{\textit{Reasoning and verification VLM}} \\
        GPT-6.1-sol  & \textbf{65.98} & \textbf{68.94} \\
        GPT-6 & 65.59 & 68.79 \\
        GPT-5.6 & 65.63 & 68.48 \\
        Qwen3.5-9B~\cite{qwen3.5} & 65.22 & 67.84 \\
        \bottomrule
    \end{tabular}
\end{minipage}
\vspace{-7pt}
\end{table}

\paragraph{Performance across refinement loops.}
\figref{fig:ablation_round_depth} shows how Mask Match mIoU changes
as refinement proceeds. After each loop, we evaluate the candidates
generated so far. 
mIoU increases from 67.60 at initialization to 68.94 after Loop 6.
 As the agent inspects and revises the segmentation over successive loops, the candidate segmentations achieve higher mIoU. This supports our intuition that PartAgent can progressively improve segmentation by refinement.

\paragraph{Qualitative loop evolution.}
\figref{fig:clr_evolution} shows how the segmentation changes over successive refinement loops.
In Loop 1, the agent groups the cable and its connector into one part,
while the control panel remains fragmented.
In Loop 2, the agent examines the updated segmentation and merges
fragments within the enclosure and within the control panel, while
keeping the cable, handpiece, and holder separate. The zoomed-in views show these local
corrections across loops.

\section{Conclusion}
\label{sec:conclusion}
We presented \textbf{PartAgent}, a training-free agentic framework for
3D part segmentation using frozen 2D and 3D experts. It uses multiview
semantic predictions to refine 3D expert regions and group them into
an initial segmentation. The agent then examines the result, decides
what to revise, and checks proposed changes. Feedback from these checks
guides its next action. 
PartAgent achieves 68.41 Mask Match mIoU on PartObjaverse-Tiny-WT,
exceeding the strongest competing method by 8.53 points. Across
3DCoMPaT200, PartNeXt, PartNetE, and PartObjaverse-Tiny, it achieves
the highest mIoU on each dataset, with a four-dataset mean of 69.36,
7.28 points above the best competing method's mean score.
{
    \small
    \bibliographystyle{ieeenat_fullname}
    \bibliography{main}
}

\clearpage
\appendix
\onecolumn
\setcounter{figure}{0}
\setcounter{table}{0}
\renewcommand{\thefigure}{S\arabic{figure}}
\renewcommand{\thetable}{S\arabic{table}}
\renewcommand{\theHfigure}{supp.\arabic{figure}}
\renewcommand{\theHtable}{supp.\arabic{table}}

\begin{center}
  {\Large\bfseries Supplemental Material\par}
\end{center}

\section{Ablations on the Remaining Objects}
\label{sec:supp_core_contributions_matched}
To limit API costs, we conduct the ablations in the main paper on 100 randomly sampled meshes from PartObjaverse-Tiny-WT~\cite{ma2025p3sam}. Here we repeat the ablations of our core modules, SSEI and CLR, on the remaining 100 meshes to see whether the same findings hold.
As shown in~\tabref{tab:supp_core_contributions_matched}, adding SSEI to the 3D expert improves Partition Match and Mask Match on both sets. Adding CLR on top of SSEI improves both metrics further. The improvements show that our
findings also hold on the remaining objects.

\begin{table}[!htbp]
\centering
\begin{minipage}{0.78\linewidth}
\centering\small
\setlength{\tabcolsep}{6pt}
\renewcommand{\arraystretch}{1.15}
\caption{\textbf{Ablations of PartAgent’s core modules~(SSEI and CLR) on PartObjaverse-Tiny-WT.}
Partition and Mask denote Partition Match and Mask Match mIoU
(\%, $\uparrow$).}
\label{tab:supp_core_contributions_matched}
\begin{tabularx}{\linewidth}{@{}l*{2}{>{\centering\arraybackslash}X}@{\hspace{18pt}}*{2}{>{\centering\arraybackslash}X}@{}}
\toprule
& \multicolumn{2}{c}{Original 100}
& \multicolumn{2}{c}{Remaining 100} \\
\cmidrule(lr){2-3}\cmidrule(lr){4-5}
Variant & Partition & Mask & Partition & Mask \\
\midrule
3D Expert only & 58.51 & 58.51 & 58.84 & 58.84 \\
3D Expert + CLR & 62.72 & 64.37 & 62.56 & 63.66 \\
SSEI only & 65.11 & 67.60 & 63.73 & 66.77 \\
Full PartAgent & \textbf{65.98} & \textbf{68.94} & \textbf{64.73} & \textbf{67.88} \\
\bottomrule
\end{tabularx}
\end{minipage}
\end{table}

\FloatBarrier
\vspace{24pt}
\section{More Qualitative Comparisons}
\label{sec:supp_qualitative}

\begin{figure}[H]
    \centering
    \includegraphics[width=\linewidth]{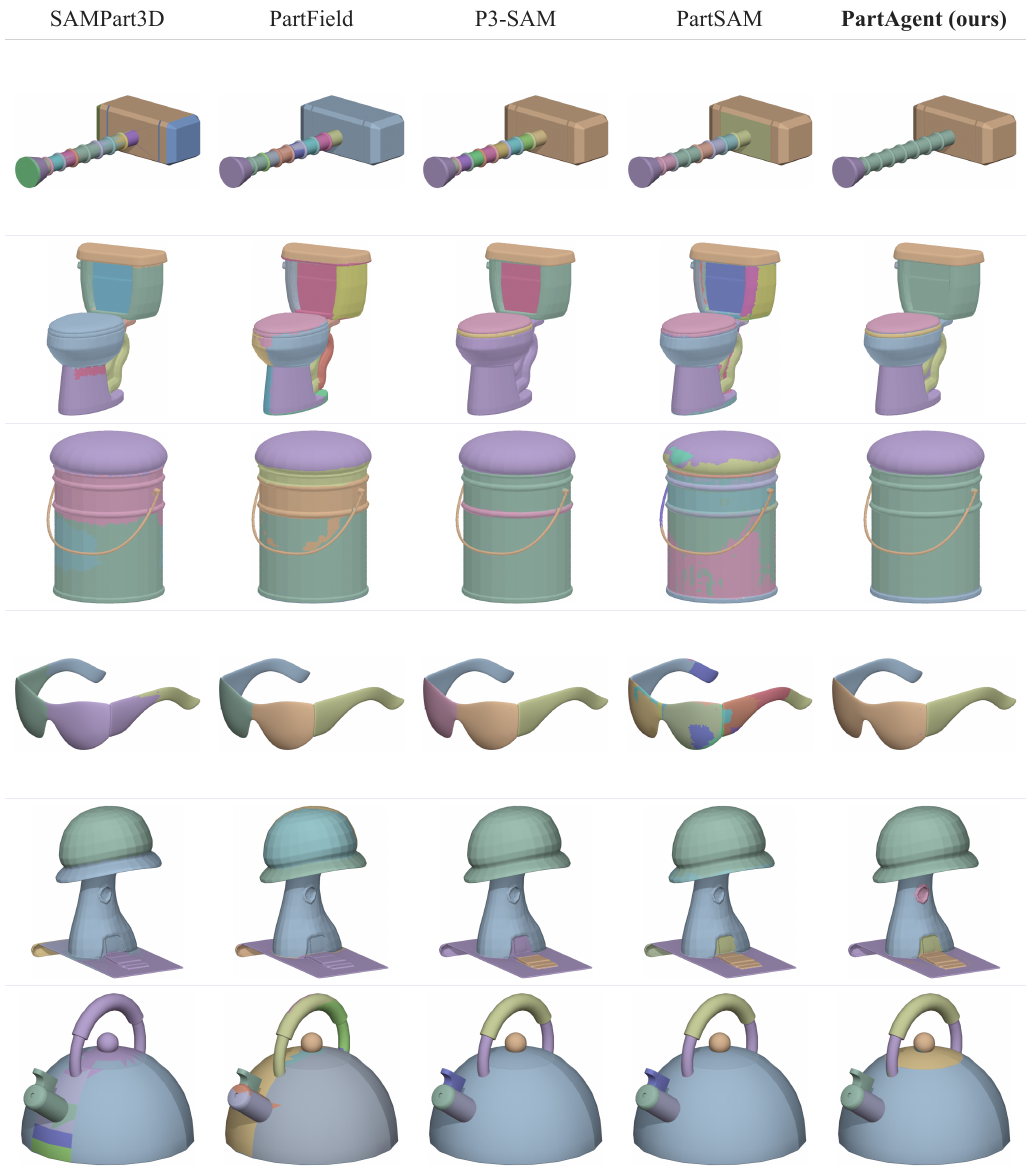}
    \caption{\textbf{Qualitative comparisons (1/2).}
    PartAgent reduces fragmentation on continuous surfaces while keeping adjacent parts separate.}
    \label{fig:qualitative_comparison}
\end{figure}

\begin{figure}[H]
    \centering
    \includegraphics[width=\linewidth]{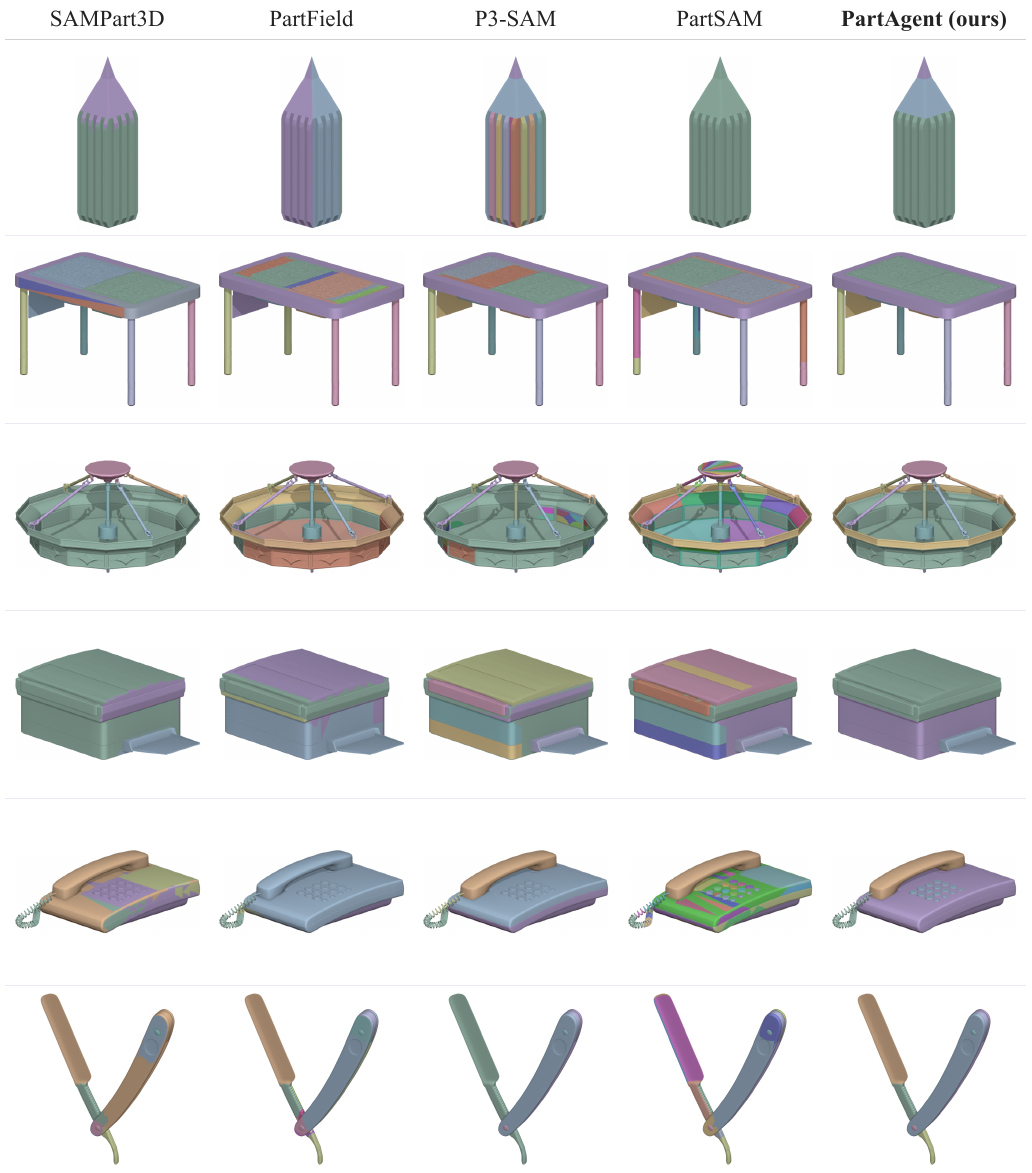}
    \caption{\textbf{Qualitative comparisons (2/2).}
    PartAgent reduces fragmentation on continuous surfaces while keeping adjacent parts separate.}
    \label{fig:qualitative_supp_cross}
    \label{fig:qualitative_supp_objects}
\end{figure}

\figref{fig:qualitative_comparison} and \figref{fig:qualitative_supp_objects}
show examples comparing PartAgent with the competitive methods, i.e., 
SAMPart3D~\cite{yang2024sampart3d}, PartField~\cite{liu2025partfield},
P3-SAM~\cite{ma2025p3sam}, and PartSAM~\cite{zhu2026partsam}.
Colors distinguish predicted parts.

\clearpage

\clearpage

\section{Additional Ablation Analyses}
\label{sec:supp_ablations}
\subsection{Qualitative Loop Evolution}
\label{sec:supp_loop_evolution}
\begin{figure}[H]
    \centering
    \includegraphics[page=1,width=\textwidth]{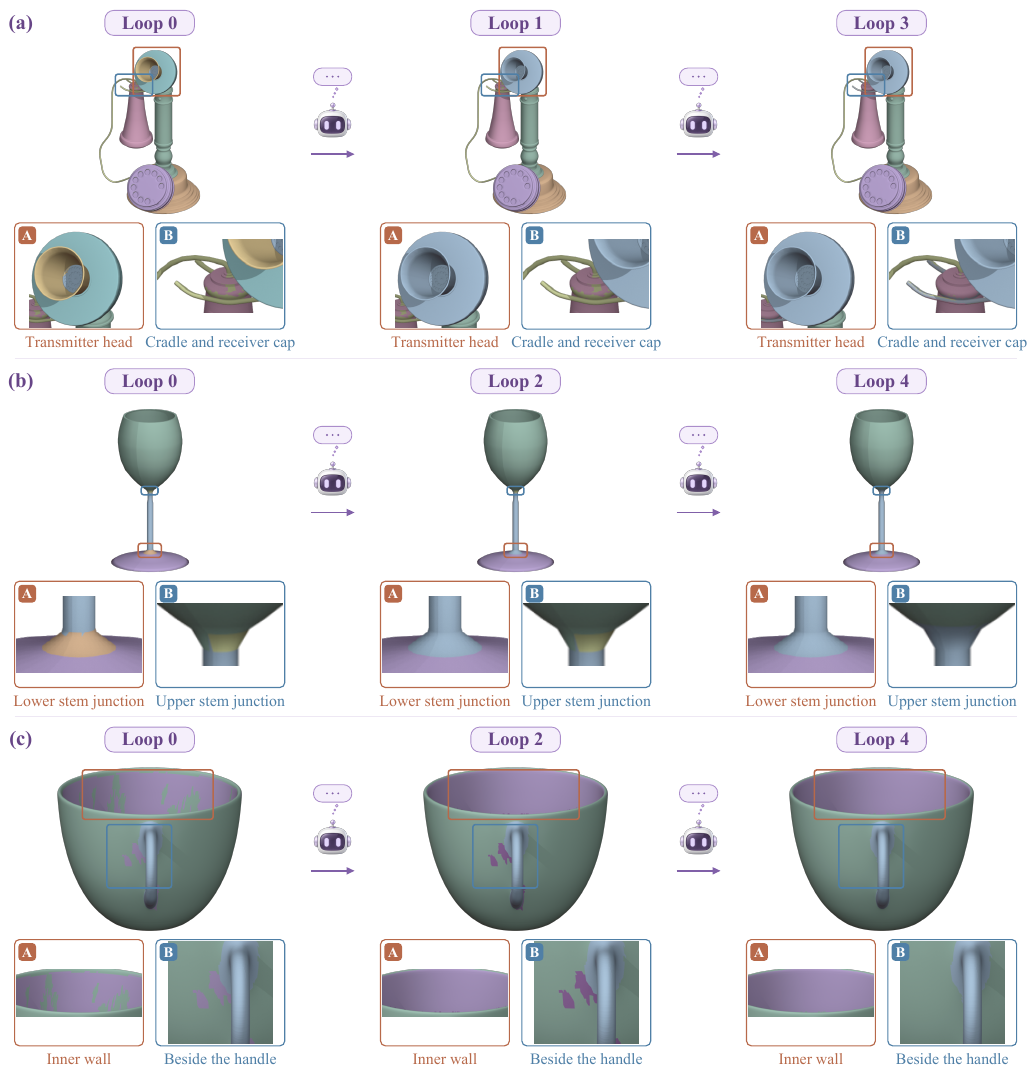}
    \caption{\textbf{Qualitative closed-loop evolution (1/2).}
    Observation-Driven Closed-Loop Partition Refinement progressively improves part segmentation over multiple loops.}
    \label{fig:supp_loop_evolution_a}
\end{figure}

\begin{figure*}[p]
    \centering
    \includegraphics[page=2,width=\textwidth]{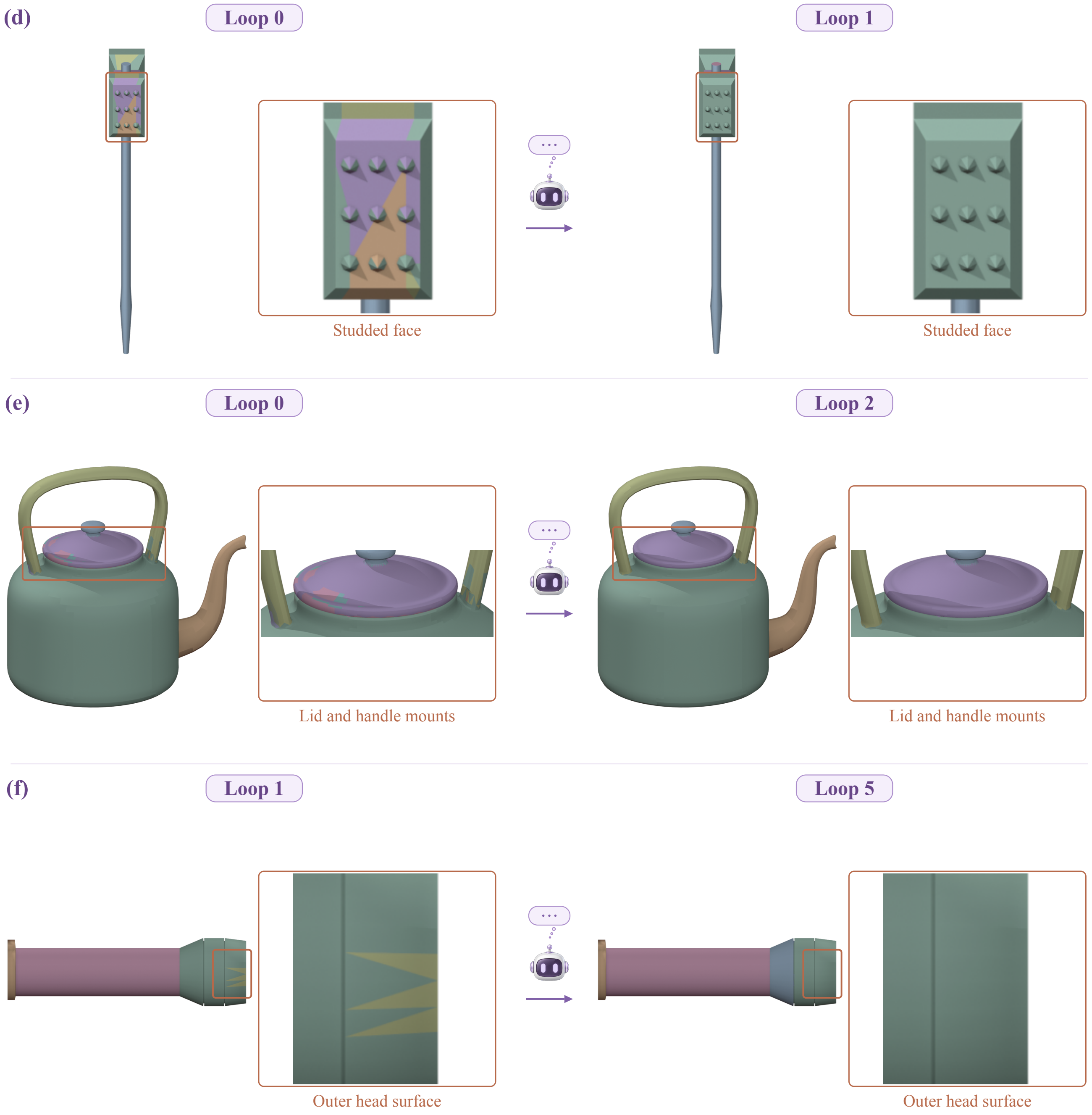}
    \caption{\textbf{Qualitative closed-loop evolution (2/2).}
    Observation-Driven Closed-Loop Partition Refinement progressively improves part segmentation over multiple loops.}
    \label{fig:supp_loop_evolution_b}
\end{figure*}

\figref{fig:supp_loop_evolution_a} and \figref{fig:supp_loop_evolution_b}
show examples of refinement. Loop~0 denotes the SSEI initialization,
and later numbers denote the  loop rounds.
The insets use the same camera, location, and magnification at each loop.

\clearpage

\providecommand{\abltbd}{\textemdash}
\providecommand{\ablna}{\textit{n/a}}

\subsection{CLR Action Ablations}
\label{sec:supp_clr_actions}

\begin{figure}[H]
    \centering
    \includegraphics[width=\linewidth]{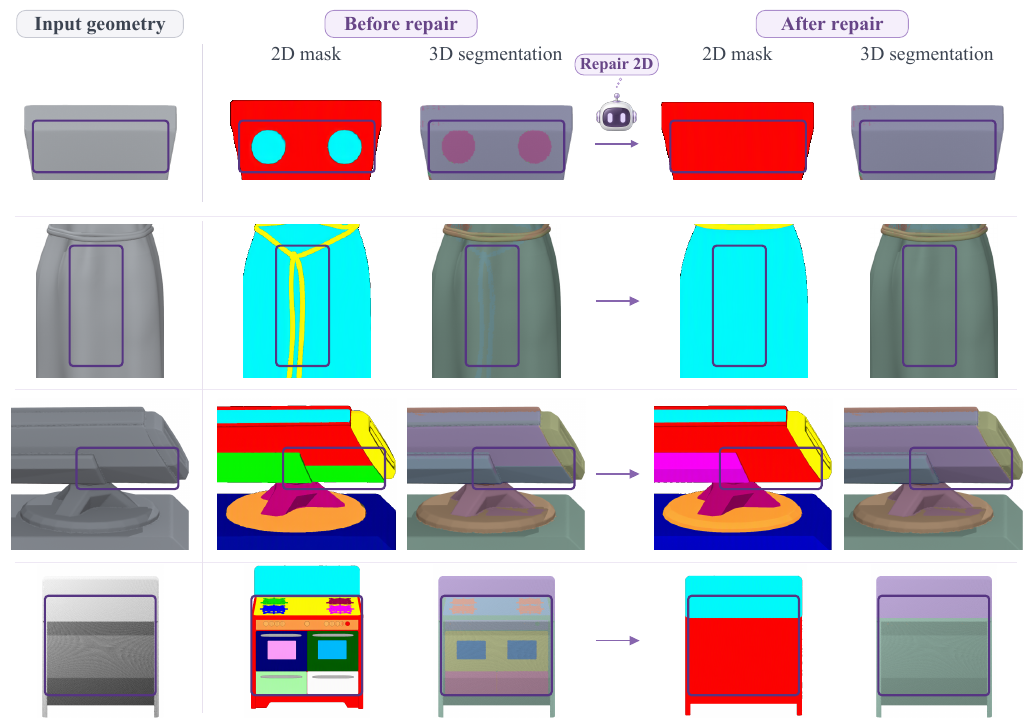}
    \caption{\textbf{Repair 2D examples.} Each row shows the same cropped view
    before and after repair. Repair~2D removes hallucinated regions and
    corrects other errors in the 2D expert's masks, improving the
    corresponding 3D segmentation.}
    \label{fig:supp_repair2d_robot}
\end{figure}

We further ablate `Compose', `Repair 2D', and `Reselect' to examine how
these actions contribute to closed-loop refinement. As shown in
\tabref{tab:supp_clr_actions}, removing `Compose' lowers Partition and
Mask by 0.86 and 1.16 points, respectively. 
Removing `Reselect' lowers them by 0.25
and 0.29 points.
And removing `Repair 2D' lowers
them by 0.02 and 0.09 points. Repair~2D yields small average gains but helps prevent an occasional
failure that matters in practice. When the 2D expert hallucinates a part,
the incorrect mask can introduce a false part into the 3D segmentation.
\figref{fig:supp_repair2d_robot} provides qualitative examples of how
repairing the 2D masks removes false regions and corrects misplaced boundaries in the 3D segmentation. This matters in practice when a part hallucinated by the 2D expert would otherwise appear as a separate part in the final result.

\begin{table}[H]
\centering\small
\setlength{\tabcolsep}{2pt}
\caption{\textbf{Observation-Driven Closed-Loop Partition Refinement.}
All variants start from the same SSEI initialization and use the
Original 100 meshes in~\tabref{tab:supp_core_contributions_matched}. Partition and Mask
report Partition Match and Mask Match mIoU (\%, $\uparrow$), respectively.
}
\label{tab:supp_clr_actions}
\begin{tabular*}{\columnwidth}{@{\extracolsep{\fill}}cccrr@{}}
\toprule
Compose & Repair 2D & Reselect & Partition & Mask \\
\midrule
-- & $\checkmark$ & $\checkmark$ & 65.12 & 67.78 \\
$\checkmark$ & -- & $\checkmark$ & 65.96 & 68.85 \\
$\checkmark$ & $\checkmark$ & -- & 65.73 & 68.65 \\
\midrule
$\checkmark$ & $\checkmark$ & $\checkmark$ & \textbf{65.98} & \textbf{68.94} \\
\bottomrule
\end{tabular*}
\end{table}

\FloatBarrier

\FloatBarrier

\end{document}